\documentclass{article}

\usepackage[accepted]{icml2026}
\setcitestyle{numbers}
\usepackage{amsmath,amssymb,amsthm,mathtools}
\usepackage{booktabs}
\usepackage{microtype, hyperref}
\usepackage{graphicx}
\usepackage{subcaption}
\usepackage{url}
\usepackage{xcolor}
\usepackage{tikz}
\usepackage{tabularx}
\usetikzlibrary{arrows.meta,positioning,fit,calc,decorations.pathreplacing}
\usepackage{enumitem}
\usepackage{pgfplots}
\pgfplotsset{compat=1.18}
\usepackage[capitalize,noabbrev]{cleveref}

\newcommand{\X}{\mathcal{X}}
\newcommand{\R}{\mathbb{R}}
\newcommand{\sgerr}{\mathcal{E}_{\mathrm{sg}}}
\newcommand{\lpred}{\mathcal{L}_{\mathrm{pred}}}
\newcommand{\lsg}{\mathcal{L}_{\mathrm{sg}}}
\newcommand{\G}{G_\theta}

\newtheorem{definition}{Definition}

\begin{document}

\twocolumn[
  \icmltitle{Semigroup Consistency as a Diagnostic for Learned Physics Simulators}

  \begin{icmlauthorlist}
    \icmlauthor{Lennon J. Shikhman}{gatech}
  \end{icmlauthorlist}

  \icmlaffiliation{gatech}{Georgia Institute of Technology}
  \icmlcorrespondingauthor{Lennon J. Shikhman}{lj@shikhman.net}

  \icmlkeywords{
    Scientific Machine Learning,
    Partial Differential Equations,
    Neural Operators,
    Learned Physics Simulators,
    Semigroup Consistency,
    Time-Evolution Models,
    Rollout Stability,
    Model Evaluation
  }

  \vskip 0.3in
]

\printAffiliationsAndNotice{}

\begin{abstract}
Learned physics simulators are often evaluated by one-step or short-horizon prediction error, but these metrics can miss failures in temporal composition and long-horizon rollout. For autonomous, state-complete systems, exact solution maps satisfy a semigroup law: direct evolution over \(s+t\) should agree with evolution over \(s\) followed by \(t\). We propose normalized semigroup error as a post hoc, model-agnostic diagnostic comparing these direct and composed learned predictions. On one-dimensional heat and Burgers dynamics with time-conditioned ConvNet and FNO baselines, semigroup error is positively associated with rollout degradation, with trajectory-level Spearman correlation \(\rho=0.635\) and \(95\%\) CI \([0.621,0.649]\). Semigroup regularization has mixed effects, supporting semigroup consistency primarily as an evaluation diagnostic rather than a universally beneficial training objective.
\end{abstract}

\section{Introduction}
\label{sec:introduction}

Learned physics simulators provide fast surrogates for PDE-governed systems such as diffusion, fluids, and waves. Neural operators and related models are especially natural for this setting because they learn maps between function spaces rather than finite-dimensional vectors \citep{lu2021deeponet,li2021fourier}. This operator-learning viewpoint has become a standard framework for PDE surrogate modeling \citep{kovachki2023neuraloperator}. Related operator-learning models include multipole and graph constructions \citep{li2020mgno,li2020gno}. Broader neural-operator formulations further develop this direction for scientific simulation \citep{azizzadenesheli2024neuraloperators}. Recent work extends these ideas with physics-informed losses and physics-informed operator learning \citep{raissi2019pinn,karniadakis2021piml}. Physics-informed neural operators provide another route to combining operator learning with equation structure \citep{li2023pino}. Other extensions include transformer-style architectures and graph operators \citep{li2023transformer,shih2025tno}. Large-scale learned simulation pipelines further demonstrate the promise of learned surrogates for physical systems \citep{Kochkov2021mlcfd}.

Standard evaluation often emphasizes one-step or short-horizon prediction error. This is useful but insufficient: models can perform well locally while failing under rollout, time-step shift, boundary-condition shift, or broader distribution shift. These issues appear in physics-informed learning and autoregressive PDE benchmarks \citep{Krishnapriyan2021PINNFailure,koehler2024apebenchbenchmarkautoregressiveneural}. They also arise in large PDE datasets and out-of-distribution studies of learned surrogates \citep{ohana2025welllargescalecollectiondiverse,nguyen2026outofdistributiongeneralizationdeeplearningsurrogates}. Recent work on neural physics solvers likewise emphasizes out-of-distribution generalization failures \citep{wei2026outofdistributiongeneralizationneuralphysics}. Recent failure-mode studies similarly show that in-distribution accuracy can obscure rollout and regime-shift errors \citep{shikhman2026diagnosing}. A simulator should not merely predict the next frame; it should define a coherent time-evolution rule.

For autonomous well-posed systems, the exact solution maps \(\{S_t\}_{t\geq 0}\) satisfy the semigroup law \citep{pazy2012semigroups,evans2010partial}:
\begin{equation}
    S_{t+s} = S_t \circ S_s, \qquad S_0 = I.
    \label{eq:intro-semigroup-law}
\end{equation}
Thus, direct evolution by \(t+s\) should agree with evolution by \(s\) followed by \(t\). Related compositional structure appears in flow-map learning and neural ODEs \citep{chen2018neuralode,qin2019data}. Data-driven PDE learning in modal space also studies evolution-map structure \citep{wu2020data}. Similar ideas appear in Koopman modeling and variational Markov models \citep{kutz2016koopmantheorypartialdifferential,mardt2018vampnets}. Semigroup-aware learning makes this structure explicit in learned evolution models \citep{chen2023deeppsg}. Similar stability concerns arise in learned dissipative dynamics for chaotic systems \citep{li2022learningdissipativedynamicschaotic}.

This paper proposes \emph{semigroup consistency} as a diagnostic for learned physics simulators. For a learned model \(G_\theta(u,t)\approx S_tu\), we compare the direct prediction \(G_\theta(u,s+t)\) with the composed prediction \(G_\theta(G_\theta(u,s),t)\). Large disagreement indicates that the learned map is not temporally consistent, even if supervised error is small. This is aligned with concerns that neural operators can learn brittle, regime-specific, or boundary-indexed solution families rather than robust physical evolution rules \citep{shikhman2026one,mousavi2026imposingboundaryconditionsneural}. It also connects to broader concerns in domain generalization \citep{Zhou_2022}.


\begin{figure}[t]
\centering
\scriptsize
\begin{tikzpicture}[
    >=Latex,
    node distance=0.38cm,
    stageA/.style={
        draw=blue!60!black,
        rounded corners,
        very thick,
        align=center,
        fill=blue!8,
        text width=0.54\columnwidth,
        minimum height=0.52cm,
        inner sep=2.8pt
    },
    stageB/.style={
        draw=teal!60!black,
        rounded corners,
        very thick,
        align=center,
        fill=teal!8,
        text width=0.54\columnwidth,
        minimum height=0.52cm,
        inner sep=2.8pt
    },
    stageC/.style={
        draw=orange!75!black,
        rounded corners,
        very thick,
        align=center,
        fill=orange!10,
        text width=0.54\columnwidth,
        minimum height=0.85cm,
        inner sep=3.0pt
    },
    stageD/.style={
        draw=red!65!black,
        rounded corners,
        very thick,
        align=center,
        fill=red!8,
        text width=0.54\columnwidth,
        minimum height=0.52cm,
        inner sep=2.4pt
    },
    stageE/.style={
        draw=purple!60!black,
        rounded corners,
        very thick,
        align=center,
        fill=purple!8,
        text width=0.54\columnwidth,
        minimum height=0.52cm,
        inner sep=2.8pt
    }
]

\node[stageA] (data) {
\textbf{1. PDE trajectories}\\
\(u(t_0),\ldots,u(t_T)\)
};

\node[stageB, below=of data] (model) {
\textbf{2. Learn simulator}\\
\(G_\theta(u,\Delta t)\approx S_{\Delta t}u\)
};

\node[stageC, below=of model] (paths) {
\textbf{3. Compare two learned paths}\\[2pt]
\(\hat u_{\rm dir}=G_\theta(u,s+t)\)\\[3pt]
\(\hat u_{\rm comp}=G_\theta(G_\theta(u,s),t)\)
};

\node[stageD, below=of paths] (sg) {
\textbf{4. Semigroup error}\\[-1pt]
\(
\|\hat u_{\rm comp}-\hat u_{\rm dir}\|_2/
(\|\hat u_{\rm dir}\|_2+\varepsilon)
\)
};

\node[stageE, below=of sg] (analysis) {
\textbf{5. Diagnostic test}\\
compare semigroup error with rollout error
};

\draw[->, very thick, blue!70!black] (data) -- (model);
\draw[->, very thick, teal!70!black] (model) -- (paths);
\draw[->, very thick, orange!85!black] (paths) -- (sg);
\draw[->, very thick, red!70!black] (sg) -- (analysis);

\end{tikzpicture}
\caption{Evaluation pipeline for semigroup consistency. A learned simulator is trained on PDE trajectories, then evaluated by comparing direct and composed learned evolution on held-out states.}
\label{fig:intro-pipeline}
\end{figure}
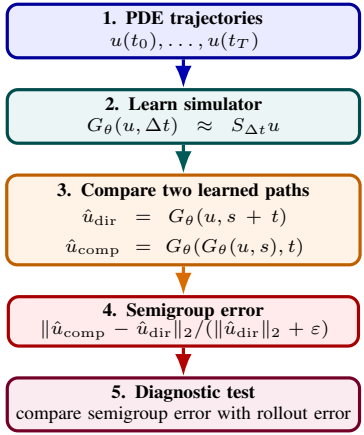

Our claim is that semigroup error complements standard prediction metrics. It can expose rollout instability, time-step extrapolation failure, fixed-step overfitting, and disagreement between direct and autoregressive simulation. Since it requires no architectural change or retraining, it can be applied to neural operators, autoregressive simulators, and continuous-time models \citep{raissi2020hiddenfluid,brunton2020fluidml}. It also applies to other learned surrogates used in physical forecasting \citep{kurth2023fourcastnet}. Empirically, we find that semigroup error is positively associated with rollout degradation on heat and Burgers dynamics, while semigroup regularization has mixed effects and does not yield a reliable aggregate rollout improvement.

Our contributions are:
\begin{enumerate}
    \item We formulate semigroup consistency as a post hoc, model-agnostic diagnostic for learned physics simulators.
    \item We define a normalized semigroup error comparing direct and composed learned evolution without requiring additional ground-truth solves.
    \item We show that semigroup error is positively associated with rollout degradation and can reveal failures hidden by one-step prediction metrics.
    \item We clarify the distinction between semigroup consistency as an evaluation diagnostic and semigroup consistency as a training regularizer.
\end{enumerate}

\section{Background and Related Work}
\label{sec:background}

\subsection{Evolution Equations and Semigroups}
\label{subsec:semigroups}

Many physical simulators model autonomous, state-complete evolution. Here, autonomous means that the dynamics do not depend explicitly on absolute time, while state-complete means that the current state contains the information needed to determine future evolution. This is closely related to the Markov property, but distinct from autonomy: a system can be Markovian and non-autonomous, or autonomous but non-Markovian if the chosen state omits relevant history. Let \(\mathcal{X}\) be a state space and let \(S_t:\mathcal{X}\to\mathcal{X}\) map an initial state \(u_0\) to its state after elapsed time \(t\). Under standard well-posedness assumptions, autonomous solution maps form a one-parameter semigroup \citep{pazy2012semigroups,evans2010partial},
\begin{equation}
    S_{t+s} = S_t \circ S_s, \qquad S_0 = I.
    \label{eq:semigroup-law}
\end{equation}
This identity states that evolving for \(s\) and then \(t\) must match evolving directly for \(s+t\). Thus, the semigroup view is not only functional-analytic; it is a structural condition for coherent prediction across temporal subdivisions, linking local differential descriptions to global evolution maps through the semigroup generator \citep{pazy2012semigroups,evans2010partial}.

This operator-level view also appears in data-driven dynamics. Flow-map learning directly approximates evolution maps from trajectories \citep{qin2019data,wu2020data}. Neural ODEs define continuous-time predictions through learned differential equations \citep{chen2018neuralode}. Koopman and variational Markov methods study learned evolution through composable operators \citep{kutz2016koopmantheorypartialdifferential,mardt2018vampnets}. Learned dissipative dynamics similarly emphasize long-time stability and composition effects \citep{li2022learningdissipativedynamicschaotic}. Deep-OSG is closest to our setting because it explicitly studies operator learning with semigroup structure and temporal partition consistency \citep{chen2023deeppsg}. Prior work uses semigroup structure to design learned evolution models \citep{chen2023deeppsg}; we instead study semigroup consistency as a post hoc diagnostic for rollout and time-step failures.

\subsection{Learned Physics Simulators}
\label{subsec:learned-simulators}

Learned physics simulators replace or accelerate numerical solves with data-driven surrogates. Neural operators are central because they learn maps between function spaces rather than fixed finite-dimensional discretizations \citep{lu2021deeponet,li2021fourier}. General neural operator theory formalizes this function-space perspective for PDEs \citep{kovachki2023neuraloperator,azizzadenesheli2024neuraloperators}. DeepONet introduced branch--trunk operator approximation \citep{lu2021deeponet}, while the Fourier Neural Operator uses global spectral mixing and is a standard baseline for parametric PDEs \citep{li2021fourier}. Later variants extend operator learning through graph and multipole constructions \citep{li2020gno,li2020mgno}. Convolutional, transformer, and finite-regularity constructions further broaden neural-operator architectures \citep{raonic2023convolutional,li2023transformer,shih2025tno}. Physics-informed methods incorporate equation residuals or physical structure into training \citep{raissi2019pinn,karniadakis2021piml}. Physics-informed neural operators extend this idea to operator learning \citep{li2023pino,li2023physicsinformedneuraloperatorlearning}.

In fluid and physical simulation, learned models support surrogate CFD and flow reconstruction \citep{raissi2020hiddenfluid,brunton2020fluidml}. They also support machine-learning-accelerated CFD and turbulence emulation \citep{Kochkov2021mlcfd,stachenfeld2022turbulencesim}. Weather prediction provides another large-scale setting for learned physical surrogates \citep{kurth2023fourcastnet}. These uses make long-horizon coherence important: a model may perform well at a fixed training interval but fail under rollout, unseen time steps, or shifted physical parameters. Time-conditioned and variable-time models address this by learning \(G_\theta(u,\Delta t)\approx S_{\Delta t}u\) rather than a single fixed-step map \citep{qin2019data,wu2020data}. Semigroup-aware operator learning also directly targets consistency across temporal partitions \citep{chen2023deeppsg}. This setting makes semigroup consistency directly testable by comparing direct prediction at \(s+t\) with composed prediction through \(s\) and \(t\).

\subsection{Reliability Beyond One-Step Error}
\label{subsec:reliability}

One-step or short-horizon prediction error is necessary but incomplete. Benchmarks such as PDEBench and APEBench broaden evaluation to rollout, resolution transfer, parameter shift, and perturbations \citep{takamoto2022pdebench,koehler2024apebenchbenchmarkautoregressiveneural}. The Well further expands large-scale physics simulation datasets for machine learning \citep{ohana2025welllargescalecollectiondiverse}. Recent robustness studies likewise show that in-distribution accuracy may not predict behavior under changes in PDE family, parameters, boundary or terminal conditions, resolution, or rollout horizon \citep{Zhou_2022,setinek2025simshift}. Additional studies emphasize out-of-distribution generalization failures in neural PDE and physics solvers \citep{nguyen2026outofdistributiongeneralizationdeeplearningsurrogates,wei2026outofdistributiongeneralizationneuralphysics}. Boundary and terminal conditions are particularly important because varying them can turn one operator-learning problem into a family of condition-indexed operators \citep{mousavi2026imposingboundaryconditionsneural}.

These failures motivate structural diagnostics beyond pointwise accuracy. Known issues include physics-informed optimization pathologies \citep{Krishnapriyan2021PINNFailure}. Neural operators can also exhibit spectral or discretization sensitivity \citep{kovachki2023neuraloperator,azizzadenesheli2024neuraloperators}. Autoregressive rollout can introduce additional instability \citep{li2022learningdissipativedynamicschaotic,koehler2024apebenchbenchmarkautoregressiveneural}. Semigroup error targets a complementary failure mode: disagreement between direct and composed evolution. Unlike supervised error, it does not require ground-truth solves at every composed intermediate state; unlike conservation or energy diagnostics, it is model-agnostic across autonomous evolution equations. It is not sufficient for correctness, since a wrong model can be internally consistent, but it is a lightweight test of whether a learned simulator behaves like a coherent time-evolution map. We therefore report semigroup consistency alongside one-step error, rollout error, and physics-specific checks.

\section{Problem Setup}
\label{sec:problem-setup}

\subsection{Physical Systems}
\label{subsec:systems}

We study autonomous PDEs with solution maps satisfying the semigroup law in \cref{eq:semigroup-law}. The experiments use two lightweight one-dimensional systems: the heat equation and viscous Burgers equation. The heat equation,
\begin{equation}
    \partial_t u = \nu \Delta u,
    \label{eq:heat}
\end{equation}
provides a stable dissipative test case. The viscous Burgers equation,
\begin{equation}
    \partial_t u + u\partial_x u = \nu \partial_{xx}u,
    \label{eq:burgers}
\end{equation}
adds nonlinear transport while remaining computationally simple. Both systems are evaluated with periodic boundary conditions so that the experiment isolates temporal composition rather than boundary-condition effects.

\subsection{Learned Simulator}
\label{subsec:model}

Let \(S_tu\) denote the reference solution after elapsed time \(t\). We study time-conditioned learned simulators
\begin{equation}
    \G(u,\Delta t) \approx S_{\Delta t}u,
\end{equation}
where \(\G\) is instantiated as either a time-conditioned convolutional network or a compact Fourier Neural Operator. The same learned map can be queried at different time increments, which makes it possible to compare direct evolution \(\G(u,s+t)\) with composed evolution \(\G(\G(u,s),t)\).

\subsection{Training Objective}
\label{subsec:training}

The base objective is supervised prediction against reference trajectories:
\begin{equation}
    \lpred(\theta)
    =
    \mathbb{E}_{(u,t)}
    \left[
    \left\|\G(u,t)-S_tu\right\|_2^2
    \right].
    \label{eq:prediction-loss}
\end{equation}
This measures direct accuracy but does not enforce temporal composition. For the semigroup-regularized ablation, we add
\begin{equation}
\begin{aligned}
    \lsg(\theta)
    =
    \mathbb{E}_{(u,s,t)}
    \big[
    &\|\G(\G(u,s),t) \\
    &\quad - \G(u,s+t)\|_2^2
    \big],
\end{aligned}
\label{eq:semigroup-loss}
\end{equation}
and train with
\begin{equation}
    \mathcal{L}(\theta)=\lpred(\theta)+\lambda_{\mathrm{sg}}\lsg(\theta).
    \label{eq:total-loss}
\end{equation}
The regularized variant is not the main contribution; it is included to distinguish semigroup consistency as an evaluation diagnostic from semigroup consistency as a training objective.
\section{Semigroup Consistency Diagnostic}
\label{sec:diagnostic}

\subsection{Definition}
\label{subsec:diagnostic-definition}

For an autonomous, state-complete system, temporal evolution is not an arbitrary collection of input-output maps: the maps must compose according to the semigroup law. This gives a simple theoretical test for learned simulators. If \(\G(u,t)\) approximates \(S_tu\), then the two learned paths
\[
    \G(u,s+t)
    \qquad\text{and}\qquad
    \G(\G(u,s),t)
\]
should agree. Their discrepancy measures whether the learned simulator behaves like a coherent time-evolution map.

\begin{definition}[Semigroup error]
Given a learned simulator \(\G:\X\times\R_+\to\X\), define
\begin{equation}
    \sgerr(u;s,t)
    =
    \frac{\left\|\G(\G(u,s),t)-\G(u,s+t)\right\|_2}
    {\left\|\G(u,s+t)\right\|_2 + \varepsilon}.
    \label{eq:semigroup-error}
\end{equation}
\end{definition}

The normalization makes the metric comparable across states and times with different magnitudes. The small constant \(\varepsilon>0\) prevents instability when the predicted state norm is close to zero, which can occur in dissipative systems. A small value of \(\sgerr\) does not prove physical correctness, but a large value indicates that the learned model violates a necessary composition property of autonomous evolution.

\subsection{Diagnostic Protocol}
\label{subsec:diagnostic-protocol}

Given a trained simulator, we sample held-out states \(u\) and evaluation time pairs \((s,t)\) such that \(s,t>0\) and \(s+t\) lies within the trajectory horizon. For each pair, we compute the direct prediction
\[
    \widehat{u}_{\mathrm{dir}} = \G(u,s+t)
\]
and the composed prediction
\[
    \widehat{u}_{\mathrm{comp}} = \G(\G(u,s),t).
\]
The diagnostic is the normalized discrepancy between these two predictions, averaged over states, time pairs, random seeds, systems, and model classes.

We evaluate two regimes. Seen time pairs use increments similar to those used during training. Unseen time pairs use larger or less common decompositions. This distinction matters because a model may learn accurate local interpolation without learning a consistent family of time-indexed maps. In that case, semigroup error should remain small on familiar increments but increase under unseen temporal decompositions.

We use semigroup error in two ways. First, as a post hoc diagnostic, it measures whether a trained simulator is internally consistent. Second, as an association test, it evaluates whether composition inconsistency is statistically linked to downstream failures such as rollout drift or time-step extrapolation error. We therefore compare \(\sgerr\) with one-step error and long-horizon rollout error, and report correlations between per-trajectory semigroup error and rollout degradation.

\subsection{Failure Modes}
\label{subsec:failure-modes}

Semigroup error is designed to flag failure modes suggested by the structure of autonomous evolution. The guiding principle is that if a learned model does not approximately satisfy a necessary composition law, then its repeated use as a simulator is suspect even when its local prediction error is small.

\paragraph{Fixed-step overfitting.}
A model trained mainly at one time increment can learn a good approximation to \(S_{\Delta t}\) without learning the full family \(\{S_t\}_{t\geq 0}\). Theoretical consistency requires \(S_{2\Delta t}=S_{\Delta t}\circ S_{\Delta t}\), and more generally \(S_{s+t}=S_t\circ S_s\). If \(\G(u,2\Delta t)\) and \(\G(\G(u,\Delta t),\Delta t)\) disagree, the model has likely learned time-step-specific interpolation rather than a coherent evolution rule.

\paragraph{Rollout drift.}
Autoregressive simulation repeatedly feeds model outputs back into the model. If direct and composed predictions disagree, then the model's own predictions move it onto states where later updates follow a different effective dynamics. This provides a theoretical reason for high semigroup error to be associated with long-horizon rollout drift.

\paragraph{Excessive damping.}
In dissipative systems, a learned model may reduce short-horizon error by over-smoothing high-frequency content. Such a model can appear stable, but it may not preserve the correct relationship between short and long time evolution. Semigroup error can detect this when repeated small-step damping does not match direct large-step damping.

\paragraph{Temporal aliasing.}
A time-conditioned model may encode different effective dynamics at different queried time increments. Then \(G_\theta(u,s+t)\) and \(G_\theta(G_\theta(u,s),t)\) behave as if they came from different learned simulators. This failure is especially likely under unseen time pairs, where the model must extrapolate its representation of elapsed time.

\paragraph{Boundary of the diagnostic.}
These failure predictions are one-way. Large semigroup error is evidence of temporal inconsistency, but small semigroup error is not a certificate of correctness. A model can be compositionally consistent and still solve the wrong equation. Semigroup error may reveal this mismatch when repeated small-step smoothing does not agree with direct large-step smoothing, although energy or spectral diagnostics are needed to identify damping errors.

\section{Experiments}
\label{sec:experiments}

\subsection{Research Questions}
\label{subsec:research-questions}

We evaluate whether semigroup consistency is a useful diagnostic for learned simulators through five questions:
\begin{enumerate}[leftmargin=*]
    \item \textbf{Correlation:} Does semigroup error correlate with long-horizon rollout error across trajectories and model variants?
    \item \textbf{Time-composition shift:} Does semigroup error increase on larger unseen composition pairs \((s,t)\)?
    \item \textbf{Regularization:} Does semigroup regularization reduce semigroup error relative to prediction-only training?
    \item \textbf{Predictive utility:} Does reduced semigroup error improve rollout error, or does it mainly improve internal consistency?
    \item \textbf{Beyond one-step error:} Can models with similar one-step error differ materially in semigroup and rollout behavior?
\end{enumerate}

\subsection{Datasets and Splits}
\label{subsec:datasets}

We use synthetic trajectories from two one-dimensional PDE systems with periodic boundary conditions: the heat equation and viscous Burgers equation. Initial conditions are sampled as random Fourier series with decaying amplitudes. Trajectories are generated on a uniform spatial grid and saved at uniformly spaced times in \([0,1]\).

The experiments use grid size \(N_x=128\), \(21\) saved time points, and \(128/32/64\) train/validation/test trajectories. The heat equation uses viscosity \(\nu=0.05\), while Burgers uses \(\nu=0.02\). Out-of-distribution tests include unseen larger composition pairs \((s,t)\), a shifted initial-condition spectrum, and a Burgers viscosity shift to \(\nu=0.03\). Splits are performed by trajectory to avoid leakage between training and testing time pairs.

\begin{table}[h]
\centering
\caption{Dataset configuration.}
\label{tab:data-config}
\footnotesize
\setlength{\tabcolsep}{3pt}
\begin{tabularx}{\columnwidth}{@{}lclX@{}}
\toprule
System & Grid & Times & Test shifts \\
\midrule
Heat, \(\nu=0.05\) &
\(128\) &
\(21\) &
unseen \((s,t)\); IC spectrum \\

Burgers, \(\nu=0.02\) &
\(128\) &
\(21\) &
unseen \((s,t)\); IC spectrum; \(\nu\to0.03\) \\
\bottomrule
\end{tabularx}
\end{table}

\subsection{Models}
\label{subsec:models}

We evaluate two compact neural simulator families. The first is a time-conditioned residual ConvNet with one-dimensional convolutional blocks and sinusoidal time embeddings. The second is a compact one-dimensional Fourier Neural Operator (FNO1D) with truncated spectral convolutions, residual \(1\times 1\) paths, and the same time-conditioning mechanism.

For each family, we train a prediction-only baseline and a semigroup-regularized variant:
\[
    \mathcal{L}
    =
    \mathcal{L}_{\mathrm{pred}}
    +
    \lambda_{\mathrm{sg}}\mathcal{L}_{\mathrm{sg}},
    \qquad
    \lambda_{\mathrm{sg}}=0.01.
\]

\begin{table}[h]
\centering
\caption{Model families.}
\label{tab:models}
\footnotesize
\setlength{\tabcolsep}{3pt}
\begin{tabularx}{\columnwidth}{@{}l l X@{}}
\toprule
Model & Time handling & Purpose \\
\midrule
TC-Conv & \(\Delta t/T\) embedding & prediction-only local baseline \\
TC-Conv + SG & same & semigroup-regularized local model \\
FNO1D & \(\Delta t/T\) embedding & operator-style baseline \\
FNO1D + SG & same & architecture-agnostic SG test \\
\bottomrule
\end{tabularx}
\end{table}

\subsection{Metrics}
\label{subsec:metrics}

All primary errors use normalized relative \(L^2\):
\[
    \mathrm{relL2}(a,b)
    =
    \frac{\|a-b\|_2}{\|b\|_2+\varepsilon}.
\]
We report one-step error, rollout AUC error, final-time rollout error, seen semigroup error, unseen semigroup error, and Spearman correlation between per-trajectory semigroup error and rollout AUC.

The semigroup error is
\[
    \mathcal{E}_{\mathrm{sg}}(u;s,t)
    =
    \frac{
    \|G_\theta(G_\theta(u,s),t)-G_\theta(u,s+t)\|_2
    }{
    \|G_\theta(u,s+t)\|_2+\varepsilon
    }.
\]
Seen semigroup error averages this quantity over training-style time pairs. Unseen semigroup error averages it over larger composition pairs.

All reported means use bootstrap \(95\%\) confidence intervals with \(2000\) resamples. We also report paired bootstrap intervals for baseline versus semigroup-regularized differences, paired Wilcoxon signed-rank tests when available, and Cohen's \(d\) effect sizes. The default run uses \(5\) random seeds.

\subsection{Main Results}
\label{subsec:main-results}

Across all held-out trajectory evaluations, systems, architectures, and training variants, unseen semigroup error is positively associated with rollout AUC error, with global Spearman correlation
\[
    \rho = 0.635, \qquad 95\%\ \mathrm{CI}=[0.621,0.649].
\]
This supports the central diagnostic claim: larger composition inconsistency is associated with worse long-horizon rollout behavior.

Table~\ref{tab:main-results} reports in-distribution means by system, model, and training variant. One-step error and rollout error are not in one-to-one correspondence. In particular, FNO1D variants generally achieve lower one-step error than TC-Conv variants, but semigroup and rollout behavior still vary across systems and regularization settings. Semigroup regularization also does not produce a reliable aggregate improvement: the paired baseline-minus-SG difference in unseen semigroup error is \(0.0017\) with CI \([-0.0123,0.0156]\), while the paired difference in rollout AUC is \(-0.0633\) with CI \([-0.2408,0.1171]\). Overall, the results support semigroup error more strongly as a diagnostic than as a universally beneficial regularizer.

\begin{table}[h]
\centering
\caption{In-distribution quantitative results. Values are means; SG denotes unseen semigroup error.}
\label{tab:main-results}
\scriptsize
\setlength{\tabcolsep}{3pt}
\renewcommand{\arraystretch}{1.08}
\begin{tabularx}{\columnwidth}{@{}lXcccc@{}}
\toprule
System & Model & 1-step & Rollout & SG & \(\rho\) \\
\midrule
Burgers & FNO1D       & 0.066 & 0.380 & 0.058 & 0.158 \\
Burgers & FNO1D+SG    & 0.062 & 0.350 & 0.055 & 0.247 \\
Burgers & TC-Conv     & 0.118 & 0.967 & 0.136 & 0.414 \\
Burgers & TC-Conv+SG  & 0.131 & 0.764 & 0.156 & 0.592 \\
Heat    & FNO1D       & 0.081 & 0.585 & 0.093 & 0.590 \\
Heat    & FNO1D+SG    & 0.079 & 0.956 & 0.093 & 0.513 \\
Heat    & TC-Conv     & 0.145 & 0.814 & 0.102 & 0.142 \\
Heat    & TC-Conv+SG  & 0.146 & 0.928 & 0.078 & 0.260 \\
\bottomrule
\end{tabularx}
\end{table}

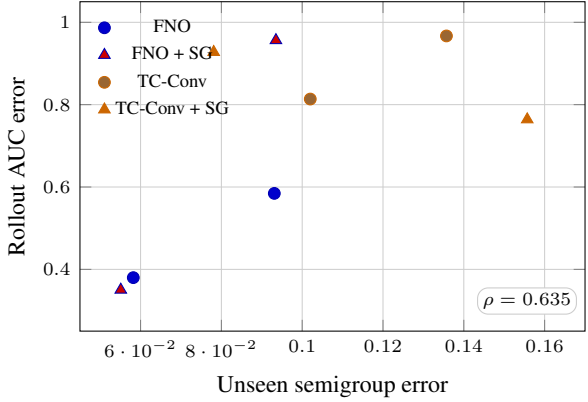
\begin{figure}[t]
\centering
\begin{tikzpicture}
\begin{axis}[
    width=\columnwidth,
    height=0.72\columnwidth,
    xlabel={Unseen semigroup error},
    ylabel={Rollout AUC error},
    xmin=0.045, xmax=0.17,
    ymin=0.25, ymax=1.05,
    grid=both,
    grid style={line width=.1pt, draw=gray!25},
    major grid style={line width=.2pt, draw=gray!40},
    tick label style={font=\scriptsize},
    label style={font=\small},
    legend style={
        font=\scriptsize,
        draw=none,
        fill=none,
        at={(0.02,0.98)},
        anchor=north west
    },
]

\addplot+[
    only marks,
    mark=*,
    mark size=2.2pt,
    color=blue!70!black
] coordinates {
    (0.0582,0.3799)
    (0.0931,0.5845)
};
\addlegendentry{FNO}

\addplot+[
    only marks,
    mark=triangle*,
    mark size=2.5pt,
    color=blue!70!black
] coordinates {
    (0.0551,0.3503)
    (0.0935,0.9563)
};
\addlegendentry{FNO + SG}

\addplot+[
    only marks,
    mark=*,
    mark size=2.2pt,
    color=orange!80!black
] coordinates {
    (0.1357,0.9669)
    (0.1020,0.8136)
};
\addlegendentry{TC-Conv}

\addplot+[
    only marks,
    mark=triangle*,
    mark size=2.5pt,
    color=orange!80!black
] coordinates {
    (0.1557,0.7640)
    (0.0781,0.9276)
};
\addlegendentry{TC-Conv + SG}

\node[
    anchor=south east,
    font=\scriptsize,
    fill=white,
    draw=gray!40,
    rounded corners,
    inner sep=2pt
] at (axis cs:0.168,0.29) {$\rho=0.635$};

\end{axis}
\end{tikzpicture}
\caption{Relationship between unseen semigroup error and rollout AUC error across systems, architectures, and training variants. Each point shows a model-level mean, while the annotated \(\rho\) reports the global trajectory-level Spearman correlation computed across all held-out evaluations.}
\label{fig:sg-vs-rollout}
\end{figure}

\subsection{Ablations}
\label{subsec:ablations}

We study two compact ablations. First, we compare semigroup error on seen and unseen composition pairs to test whether temporal inconsistency becomes more visible under time-composition shift. Averaged across all evaluated settings, unseen semigroup error exceeds seen semigroup error by \(0.0165\), indicating that larger unseen compositions expose additional inconsistency.

Second, we compare prediction-only and semigroup-regularized training to test whether directly minimizing \(\mathcal{L}_{\mathrm{sg}}\) improves downstream behavior. In aggregate, this effect is weak. The paired baseline-minus-SG difference in unseen semigroup error is \(0.0017\) with \(95\%\) CI \([-0.0123,0.0156]\), and the paired baseline-minus-SG difference in rollout AUC is \(-0.0633\) with \(95\%\) CI \([-0.2408,0.1171]\). Both intervals contain zero, so the regularization effect is not statistically reliable at the aggregate level.

\begin{figure}[t]
\centering
\begin{tikzpicture}
\begin{axis}[
    width=\columnwidth,
    height=0.70\columnwidth,
    ybar,
    bar width=5.5pt,
    ymin=0,
    ymax=0.18,
    ylabel={Semigroup error},
    symbolic x coords={FNO,FNO+SG,TC,TC+SG},
    xtick=data,
    xticklabel style={font=\scriptsize},
    tick label style={font=\scriptsize},
    label style={font=\small},
    legend style={
        font=\scriptsize,
        draw=none,
        fill=none,
        at={(0.02,0.98)},
        anchor=north west
    },
    grid=major,
    grid style={draw=gray!25},
    axis line style={draw=gray!70},
    tick style={draw=gray!70},
]

\addplot[
    fill=blue!35,
    draw=blue!70!black
] coordinates {
    (FNO,0.0901)
    (FNO+SG,0.0891)
    (TC,0.1326)
    (TC+SG,0.1263)
};
\addlegendentry{Seen}

\addplot[
    fill=orange!55,
    draw=orange!85!black
] coordinates {
    (FNO,0.0981)
    (FNO+SG,0.0894)
    (TC,0.1598)
    (TC+SG,0.1569)
};
\addlegendentry{Unseen}

\end{axis}
\end{tikzpicture}
\caption{Seen and unseen semigroup error averaged across evaluated regimes. Bars show model-level means. Unseen composition pairs generally increase semigroup error, consistent with time-composition shift.}
\label{fig:seen-vs-unseen}
\end{figure}
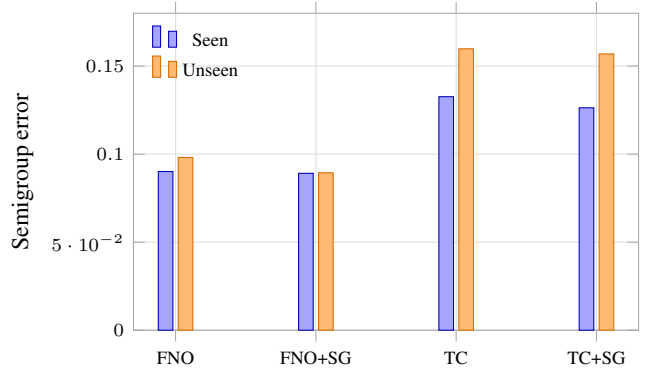

These ablations distinguish diagnostic value from regularization value. Semigroup error reveals temporal inconsistency under unseen compositions, but explicitly minimizing the semigroup penalty does not reliably improve rollout behavior in aggregate.

\section{Discussion}
\label{sec:discussion}

The results support semigroup error primarily as a diagnostic for learned physics simulators. Across systems, architectures, and variants, higher unseen semigroup error is associated with worse rollout behavior. This suggests that temporal composition provides information not captured by one-step prediction error alone. A model may approximate short-time targets well while still failing to represent a coherent family of time-indexed evolution maps.

The regularization results are more mixed. Adding \(\mathcal{L}_{\mathrm{sg}}\) does not yield a reliable aggregate improvement in either semigroup error or rollout AUC. This distinction is important: semigroup consistency can be useful for evaluation even when directly optimizing a simple semigroup penalty is not sufficient to improve learned dynamics. In this sense, the diagnostic value of semigroup error is better supported here than its use as a universal training objective.

Semigroup error should also be interpreted as a one-way warning signal. Large semigroup error indicates that direct and composed learned evolution disagree, which violates a necessary structural property of autonomous evolution. Small semigroup error, however, does not imply that the learned model is physically correct. A model can be internally consistent while solving the wrong equation, using the wrong dissipation rate, or failing under parameter and boundary shifts. Semigroup consistency is therefore best reported alongside one-step error, rollout error, and physics-specific diagnostics.

\section{Limitations}
\label{sec:limitations}

This study uses compact one-dimensional heat and Burgers systems. These settings isolate temporal composition cleanly, but they do not establish that the same diagnostic behaves identically for high-dimensional fluids, turbulent flows, complex geometries, or multi-physics simulators. Extending the evaluation to larger PDE benchmarks is an important next step.

The diagnostic is most appropriate for autonomous, state-complete systems. If the dynamics are non-autonomous, controlled, stochastic, history-dependent, or driven by changing boundary conditions, the one-parameter semigroup law is not the right object without modification. In those cases, the diagnostic should be replaced by an evolution-family or cocycle consistency condition.

Semigroup consistency is not a certificate of correctness. It measures internal temporal coherence of the learned map, not agreement with the true physical dynamics. The normalization in \(\mathcal{E}_{\mathrm{sg}}\) can also become delicate for strongly dissipative systems when state norms are small. Finally, for chaotic systems, state-space rollout error may become less meaningful at long horizons, and semigroup diagnostics should be paired with statistical or distributional measures of physical behavior.

\section{Conclusion}
\label{sec:conclusion}

We proposed semigroup consistency as a post hoc diagnostic for learned physics simulators. The diagnostic compares direct learned evolution over \(s+t\) with composed learned evolution over \(s\) followed by \(t\), producing a normalized semigroup error that can be evaluated without additional ground-truth solves. Across heat and Burgers experiments, semigroup error is positively associated with rollout degradation, supporting its use as a structural warning signal beyond one-step prediction error. Semigroup regularization has mixed effects and does not produce a reliable aggregate rollout improvement, so the evidence supports semigroup consistency more strongly as an evaluation diagnostic than as a universal training objective. Future work should test this diagnostic on higher-dimensional PDEs, more realistic physics simulators, and richer composition-aware training methods.

\section*{Impact Statement}

This work aims to improve reliability evaluation for learned physics simulators by identifying temporal inconsistency that may be missed by one-step prediction error. Semigroup error is a lightweight diagnostic, not a certification method: a model can compose consistently while still being physically wrong. In high-stakes scientific or engineering settings, it should be used alongside numerical benchmarks, uncertainty quantification, physical diagnostics, and expert review.

\newpage

\section*{Acknowledgements}

\paragraph{Reproducibility.} Code to reproduce all experiments is available at \href{https://github.com/lennonshikhman/semigroups-for-physics/}{https://github.com/lennonshikhman/semigroups-for-physics/}. \paragraph{Computational Resources.} The author gratefully acknowledges Dell Technologies, and in particular the Dell Pro Max workstation team, for providing computational resources that supported the experiments in this work. Experiments were conducted on a Dell Pro Max T2 workstation equipped with an Intel Core Ultra 9 285K processor, 128 GB of DDR5 ECC memory, and an NVIDIA RTX PRO 6000 Blackwell GPU. The views and conclusions expressed herein are those of the author and do not necessarily reflect the views of Dell Technologies.

\bibliography{references}
\bibliographystyle{icml2026}

\appendix

\section{Numerical Solvers}
\label{app:numerical-solvers}

All reference trajectories are generated before training using deterministic numerical solvers. The experiments use two one-dimensional systems: the heat equation and the viscous Burgers equation, both with periodic boundary conditions. Periodic domains simplify the solver implementation and avoid boundary-condition ambiguity, which is useful because the present work studies temporal composition rather than boundary generalization.

Initial conditions are sampled as random Fourier series with controlled smoothness. For the heat equation, trajectories are generated with viscosity \(\nu=0.05\). For viscous Burgers dynamics, trajectories are generated with viscosity \(\nu=0.02\), with an additional shifted-viscosity test regime using \(\nu=0.03\). The solvers use periodic finite-difference derivatives and a sufficiently small internal time step relative to the saved trajectory spacing. The implementation checks that generated trajectories remain finite and contain no NaNs or infinities.

In the main experiments, saved trajectories have shape
\[
    [N,T,X],
\]
where \(N\) is the number of trajectories, \(T\) is the number of saved times, and \(X\) is the spatial grid size. We use \(X=128\), \(T=21\), and \(128/32/64\) train/validation/test trajectories. Datasets are split by initial condition rather than by sampled time pairs, preventing leakage between training and testing trajectories. Shifted test regimes include unseen time-pair compositions, shifted Burgers viscosity, and shifted initial-condition spectra.

\section{Model Details}
\label{app:model-details}

The experiments use compact one-dimensional learned simulators of the form
\[
    G_\theta(u,\Delta t)\approx S_{\Delta t}u.
\]
The two model families are a time-conditioned convolutional residual network and a compact one-dimensional Fourier Neural Operator (FNO). Both models receive the current state \(u\) and a normalized time increment \(\Delta t/T\). Time conditioning is implemented using sinusoidal time embeddings injected into the model layers.

The convolutional baseline consists of one-dimensional convolutional residual blocks. The FNO baseline uses spectral convolution layers with a fixed number of retained Fourier modes, followed by pointwise channel mixing. Both architectures use residual prediction, returning the input state plus a learned update.

For each model family, we train two variants. The baseline minimizes only the supervised prediction loss,
\[
    \mathcal{L}_{\mathrm{pred}}(\theta)
    =
    \mathbb{E}_{(u,t)}
    \left[
    \|G_\theta(u,t)-S_tu\|_2^2
    \right].
\]
The semigroup-regularized variant minimizes
\[
    \mathcal{L}(\theta)
    =
    \mathcal{L}_{\mathrm{pred}}(\theta)
    +
    \lambda_{\mathrm{sg}}\mathcal{L}_{\mathrm{sg}}(\theta),
\]
where
\[
    \mathcal{L}_{\mathrm{sg}}(\theta)
    =
    \mathbb{E}_{(u,s,t)}
    \left[
    \|G_\theta(G_\theta(u,s),t)-G_\theta(u,s+t)\|_2^2
    \right].
\]
The main experiments use \(\lambda_{\mathrm{sg}}=0.01\).

All models are trained with AdamW, gradient clipping, and deterministic random seeds where possible. The reported experiments use five seeds and bootstrap \(95\%\) confidence intervals with \(2000\) resamples. The implementation automatically uses CUDA when available and otherwise falls back to CPU.

\section{Non-Autonomous Extensions}
\label{app:non-autonomous}

The semigroup identity used in the main paper applies most directly to autonomous, state-complete systems. Autonomy means that the evolution rule does not depend explicitly on absolute time. State-completeness means that the current state contains the information needed to determine future evolution. Many physical systems are instead non-autonomous or not state-complete: the dynamics may depend explicitly on time, external controls, changing boundary conditions, moving geometry, stochastic forcing, or history-dependent material response. In these cases, the one-parameter semigroup law
\[
    S_{t+s}=S_t\circ S_s
\]
is generally not the correct structure.

For non-autonomous dynamics, the appropriate object is an evolution family. Let \(S(t,r)\) denote the map that evolves a state from time \(r\) to time \(t\). The composition law becomes
\[
    S(t,r)=S(t,s)\circ S(s,r),
    \qquad r\leq s\leq t.
\]
A learned non-autonomous simulator should therefore be written as
\[
    G_\theta(u,r,t)\approx S(t,r)u,
\]
or, when controls or forcing are present,
\[
    G_\theta(u,r,t,c_{[r,t]})\approx S_c(t,r)u,
\]
where \(c_{[r,t]}\) denotes the control, forcing, or boundary information over the interval.

The corresponding diagnostic compares direct and composed evolution:
\[
    \mathcal{E}_{\mathrm{evol}}(u;r,s,t)
    =
    \frac{
    \|G_\theta(G_\theta(u,r,s),s,t)-G_\theta(u,r,t)\|_2
    }{
    \|G_\theta(u,r,t)\|_2+\varepsilon
    }.
\]
This reduces to the semigroup diagnostic in the autonomous case when \(S(t,r)=S_{t-r}\). For controlled or externally forced systems, the diagnostic must also ensure that both direct and composed predictions condition on consistent forcing information over the same interval. Otherwise, disagreement may reflect mismatched inputs rather than failure of the learned dynamics.

Thus, semigroup consistency should not be applied blindly to every simulator. It is best viewed as the autonomous case of a broader composition-consistency principle: a learned simulator should agree with itself when the same physical evolution is represented either as one interval or as a composition of subintervals.

\end{document}